\documentclass[10pt,twocolumn,letterpaper]{article}

\usepackage[pagenumbers]{cvpr} 
\usepackage{algorithm}
\usepackage{algpseudocode}
\usepackage{multirow}
\usepackage{pifont}
\usepackage{amsmath}
\usepackage{booktabs}
\usepackage{amssymb}
\usepackage{threeparttable}
\usepackage[table]{xcolor}
\definecolor{cvprblue}{rgb}{0.21,0.49,0.74}
\usepackage[pagebackref,breaklinks,colorlinks,allcolors=cvprblue]{hyperref}

\def\paperID{729} 
\def\confName{CVPR}
\def\confYear{2026}

\title{Modality-Transition Representation Learning for \\ Visible-Infrared Person Re-Identification}

\author{
Chao Yuan$^{1,*}$, Zanwu Liu$^{1,*}$, Guiwei Zhang$^{1}$, Haoxuan Xu$^{1}$, Yujian Zhao$^{1}$, Guanglin Niu$^{1}$, Bo Li$^{1,\dagger}$ \\
\textsuperscript{1} Beihang University \quad \\
{\tt\small yuanc3666@gmail.com}, {\tt\small \{beihangngl,boli\}@buaa.edu.cn} \\
}

\begin{document}
\maketitle

\begin{abstract}
Visible-infrared person re-identification (VI-ReID) technique could associate the pedestrian images across visible and infrared modalities in the practical scenarios of background illumination changes. However, a substantial gap inherently exists between these two modalities. 
Besides, existing methods primarily rely on intermediate representations to align cross-modal features of the same person. The intermediate feature representations are usually create by generating intermediate images (kind of data enhancement), or fusing intermediate features (more parameters, lack of interpretability), and they do not make good use of the intermediate features, or even the performance gained from more training data.
Thus, we propose a novel VI-ReID framework via \textbf{M}odality-\textbf{T}ransition \textbf{R}epresentation \textbf{L}earning (\textbf{MTRL}) with a middle generated image as a transmitter from visible to infrared modals, which are fully aligned with the original visible images and similar to the infrared modality. After that, using a modality-transition contrastive loss and a modality-query regularization loss for training, which could align the cross-modal features more effectively. Notably, our proposed framework \textbf{does not need any additional parameters}, which achieves the same inference speed to the backbone while improving its performance on VI-ReID task. Extensive experimental results illustrate that our model significantly and consistently outperforms existing SOTAs on three VI-ReID datasets. Code: \textit{\url{https://github.com/yuanc3/MTRL}}
\end{abstract}

\section{Introduction}

Person re-identification (ReID) aims to associate specific individuals across non-overlapping camera views~\cite{ye2021deep,zhang2021unrealperson,yuan2025neighbor,xu2025identity,11209188,liu2025try}. Previous researches mainly focus on the images captured by single visible (RGB) modal cameras~\cite{he2021transreid,tan2022dynamic,zhang2023pha}. However, the information of the same pedestrian captured by the imaging devices with both visible (RGB) and infrared (IR) modalities have obvious gap, causing dramatic performance degradation in the cross-modal scenario of the traditional ReID models. Therefore, visible-infrared person ReID (VI-ReID) is a promising technique to retrieve individuals relying on both visible and infrared modal cameras.

\begin{figure}
\centering
\includegraphics[width=0.85\linewidth]{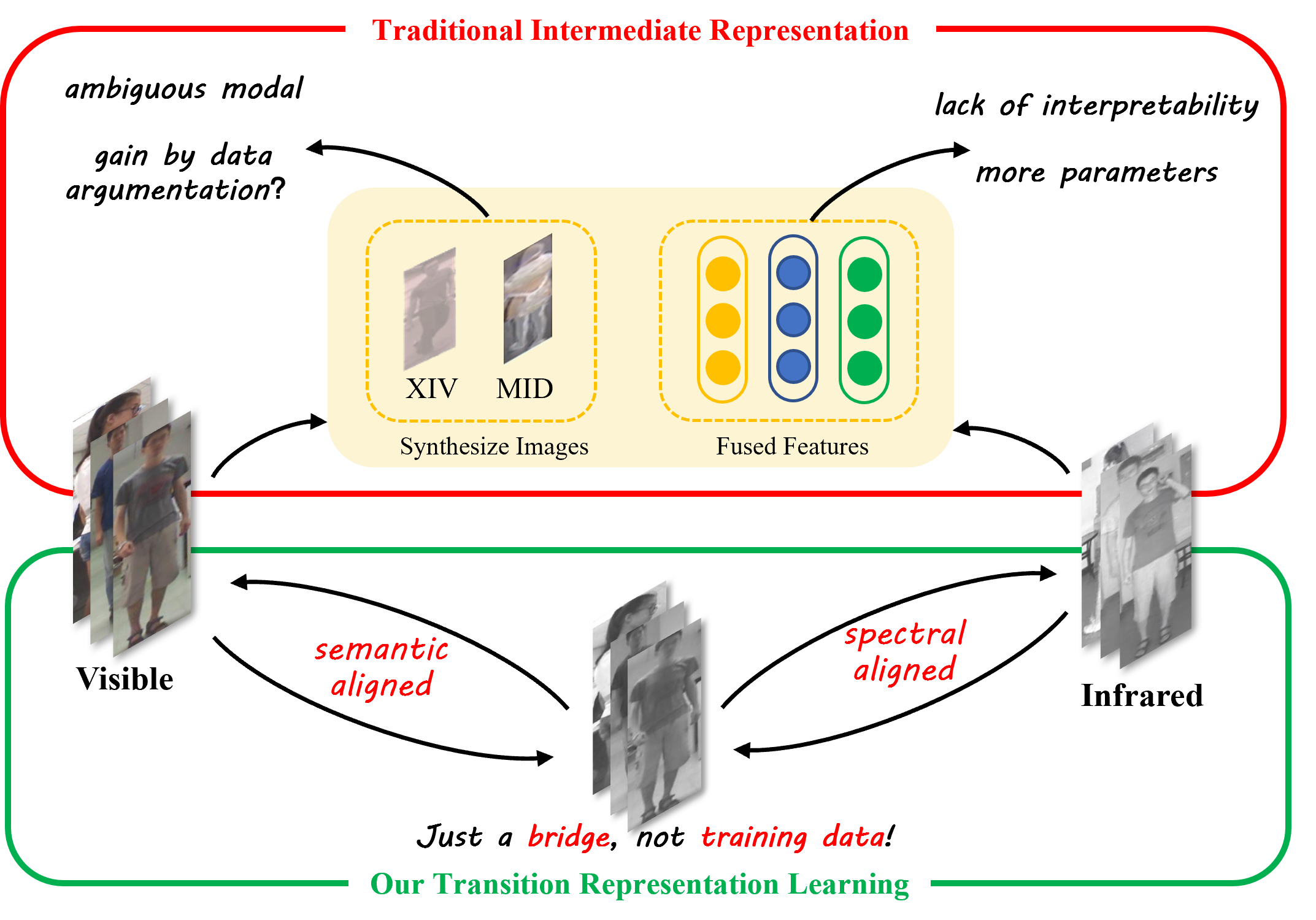}
\caption{The problems of the existing methods and our main idea of our modality-transition representation learning. XIV\cite{li2020infrared} and MID\cite{Huang2022ModalityAdaptiveMA} are some of works use generated images as intermediate, which participate training.}
\label{fig:intro}
\end{figure}

To address the significant gap between RGB and infrared modalities, many recent studies have explored the creation of intermediate modalities or intermediate features to facilitate cross-modal alignment. These approaches mainly fall into two categories: synthesizing intermediate images\cite{Huang2022ModalityAdaptiveMA,li2020infrared,agpi}, and utilizing specific modules to fuse intermediate features\cite{Zhang2023DiverseEE,zhang2022modality,10377404,jiang2022cross,zhang2022fmcnet,kim2023partmix,10205002,CAL,Ren2024ImplicitDK}. As shown in Fig.\ref{fig:intro}, the former category often generates images that exhibit a clear distribution gap from real data, requiring additional parameterized modules for separate processing, and their performance improvements may primarily stem from data augmentation. The latter relies on the model’s own learning, making it less interpretable, and necessitates extra parameterized modules, which increases inference time, while its performance gain may largely result from the increased number of parameters. 

Besides, these methods present two issues during implementation: \textbf{1)} They typically involve training with synthetic images, \textbf{implicitly performing data augmentation}. Consequently, the resulting improvements cannot be attributed to structural advantages versus training on more data. \textbf{2)} Synthetic images cannot perfectly match the distribution of real images. Directly incorporating them into training causes the model to \textbf{develop modal bias}.

To overcome these limitations, we propose a framework which does not introduce additional model parameters, and the intermediate transition image does not participate in parameter updates, ensuring that performance improvements are not attributed to the generated images.

\begin{figure*}
\centering
\includegraphics[width=\textwidth]{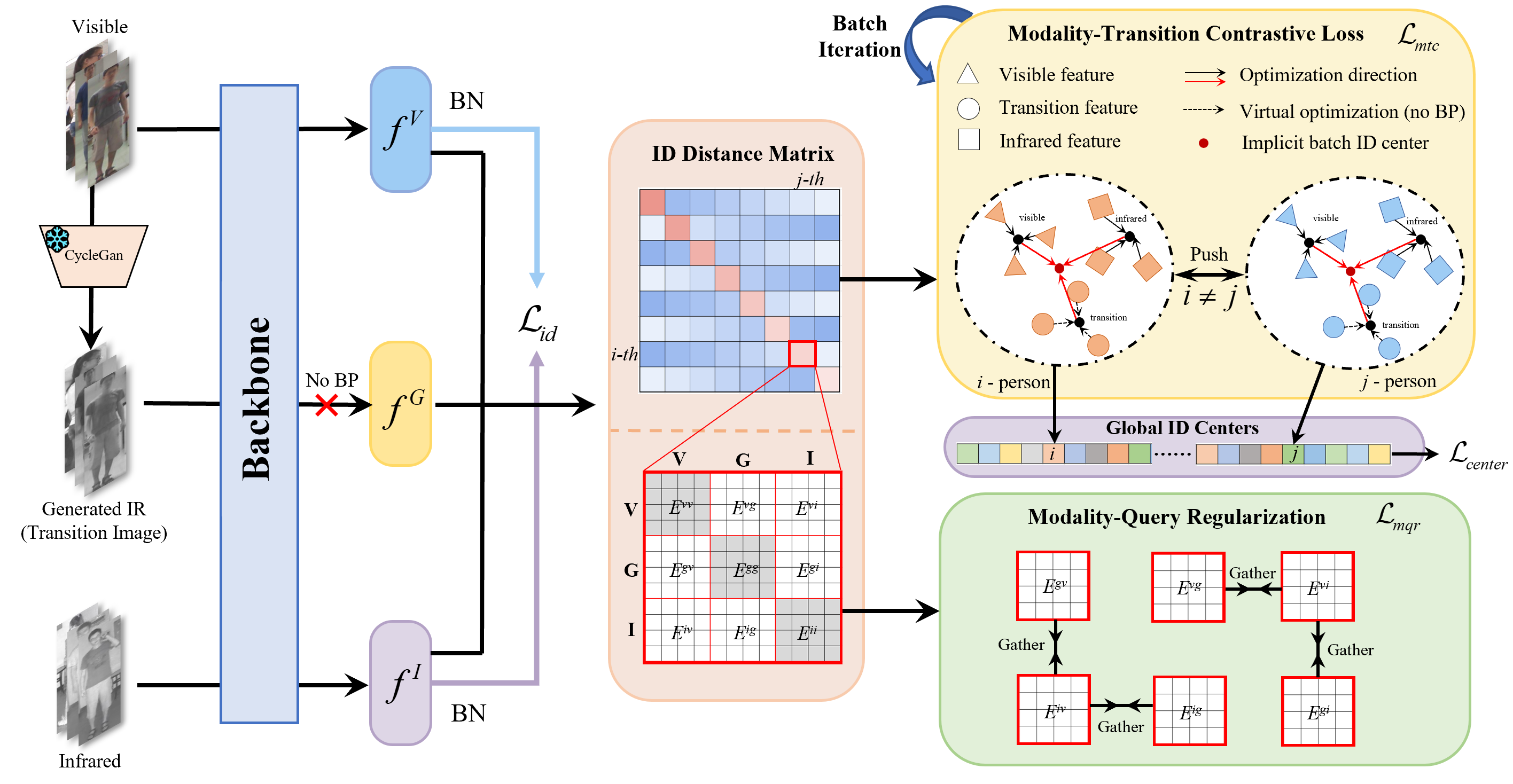}
\caption{Overview of the proposed MTRL framework for the VI-ReID task, including \ding{192} Modality-Aware Hierarchical Constraints (generative modality-transition contrastive loss $\mathcal{L}_{mtc}$ and global ID center loss $\mathcal{L}_{center}$), and \ding{193} Modality-Query Regularization $\mathcal{L}_{mqr}$.}
\label{fig:main}
\end{figure*}

Specifically, we propose \textbf{M}odality-\textbf{T}ransition \textbf{R}epresentation \textbf{L}earning, termed \textbf{MTRL}, a novel approach underpinned by three parts designs as shown in Figure~\ref{fig:main}. First, 
we propose \textbf{Modality-Aware Hierarchical Constraints} Losses consisting of a Modality-Transition Contrastive Loss and a Global Center Loss to pull the features of the same individual in different modalities closer via leveraging the generative modality-transition data. Second, based on the observation that generated images is fully aligned with visible images and closer to the infrared modality in terms of spectra, we propose a \textbf{Modality-Query Regularization} to take advantage of the variance in cross-modal queries for further cross-modal alignment. Thus, our contributions can be summarized as follows:
\begin{itemize}
    \item[$\bullet$] We propose a novel Modality-Transition Representation Learning framework for VI-ReID, which does not introduce additional parameters and inference time (\textbf{only 16\% extra forward training time}), and transition images does not participate in parameter updates, which means the improvements are derived from the framework, not data augmentation like others.

    \item[$\bullet$] We design Modality-Aware Hierarchical Constraints Loss. It utilizes the transition modality as a bridge to pull together features of the same pedestrians in different modalities under both batch and global constraints.
    
    \item[$\bullet$] We design Modality-Query Regularization Loss. It takes advantage of the transition images to bring modalities closer using cross-modality queries.

  \end{itemize}

\section{Related Works}
\subsection{Visible-Infrared Person Re-Identification}

Traditional cross-modal person re-identification (ReID) methods typically process data from two modalities separately and perform contrastive learning between them, such as \cite{Zhang2023ProtoHPEPH,9577837,10377129,Liang2021CrossModalityTW,Li2022CounterfactualIF,10814078,yuan2025poses}.

With the advancement of cross-modal ReID research, an increasing number of studies in recent years have introduced an intermediate modality or intermediate feature representation for joint learning, and this approach has been widely validated for its effectiveness. These methods generally fall into two categories:

The first way generates intermediate images to facilitate learning. For instance, XIV\cite{li2020infrared} generates X-modality images through channel transformation, MID\cite{Huang2022ModalityAdaptiveMA} synthesizes new images by combining two modality images, and $\text{AGPI}^2$ utilizes gray images as an intermediate modality.

The second way involves designing specific modules to fuse intermediate feature vectors, thereby improving cross-modal alignment. For example, MUN\cite{10377404} and FMCNet\cite{zhang2022fmcnet} introduce fusion modules to construct intermediate features, while IDKL\cite{Ren2024ImplicitDK} extracts common features for knowledge distillation.

It is worth emphasizing that our MTRL framework is free of any backbone to enhance feature alignment without extra inference overhead to improve VI-ReID performance.

\subsection{Visible-to-infrared Translation Model}

Visible and infrared image translation is an active research topic due to their advantages in detection, segmentation, and re-identification tasks~\cite{9713673,2022Visible,9362227,rs13183656}. Many systems use both visible ans infrared sensors to enhance visual signals. Basic methods for modality translation~\cite{2022Visible,2018IR2VI,2019Unpaired,2017Infrared,2021I2V} have been proposed.

For unpaired images such as visible and infrared person images, pixel-level supervised models cannot be used for training. Inspired by the unpaired generation method of CycleGAN~\cite{zhu2017unpaired}, numerous similar models~\cite{2021UNIT,2021Palette,2021High,Turner2024DenoisingDP,ho2020denoising} have emerged to address unpaired image translation tasks. This paper employs the vanilla CycleGAN to validate the effectiveness of the proposed framework.

\section{Methodology}
Common Re-ID models focus on pedestrian feature vectors extracted from images for matching. Considering that inference efficiency is important for actual person Re-ID applications, we aim to exploit the generative IR transition modality for constraint losses in the training procedure. Therefore, our model can be plugged into any Re-ID baseline framework to improve the performance of the baseline without introducing any redundant modules.

Specially, for the generated transition modality, we trained a CycleGan, which is for unpaired image generation like VI-ReID task, on the train set of VI-ReID datasets as the generator to generate images of the transition modality. Moreover, we simply use the gray images as the transition modality to show the strong ability of our framework.

\subsection{Modality-Aware Hierarchical Constraints}

Based on the generated transition modal images, we propose a \textbf{Modality-Aware Hierarchical Constraint} ($\mathcal{L}_{mhc}$) mechanism to constrain model in a training batch and a global batch between different modalities. It takes advantage of two properties of the transition modality: \ding{192} fully aligned with the visible modal images and \ding{193} more similar to the infrared image at the spectral level.

The loss $\mathcal{L}_{mhc}$ contains two tiers, namely the local-based \textbf{Modality-Transition Contrastive Loss} ($\mathcal{L}_{mtc}$) which is a batch iteration constraint, and the global-based \textbf{Center Loss} ($\mathcal{L}_{center}$) which is a global ID constraint. 
\\\textbf{ID Distance}. We define ID distance matrix \(D^{m1,m2}\) to evaluate the similarity between pair-wise features of persons in the view of various modalities. We employ a top-k distance to calculate ID distance matrix \(D^{m1,m2}\) as the following:
\begin{small} 
\begin{equation}
D^{m1,m2}_{i,j} = \frac{1}{k}\sum{ top(k, \left\lVert F^{m1}_{i,p}-F^{m2}_{j,q} \right\rVert_2)}, p\in N, q\in N 
\end{equation}
\end{small}
where $m1,m2\in \left\{V, G, I\right\}$ represent two modalities. $V$, $G$ and $I$ indicate the visible modal, the generated IR transition modality, and the original IR modal, respectively. $D^{m1,m2}_{i,j}$ denotes the distance between the $i_{th}$ and $j_{th}$ person between $m1$ and $m2$ modalities. $top(k,\cdot)$ indicate the function of selecting the maximum $k$ instances if $i= j$ while selecting the minimum $k$ instances if $i\neq j$ for $D^{m1,m2}_{i,j}$. $\boldsymbol{F}^{m1}_{i,p}$ and $\boldsymbol{F}^{m2}_{j,q}$ indicate the feature of the $i_{th}/j_{th}$ person's $p_{th}/q_{th}$ instance in the $m1/m2$ modal. $N$ denotes the positive samples' number. $||\cdot||_2$ is the Euclidean distance.

\textbf{Modality Constraint Loss}
 $\mathcal{L}_{mc}$ is a basic unit of $\mathcal{L}_{mtc}$, which is used to constrain the ID distance matrix obtained from the computation of the specified two modes, with the aim of drawing the samples with the same IDs closer while drawing the samples with different ids farther.
\begin{small}
\begin{equation}
\mathcal{L}_{mc}^{m1,m2}={\lambda}_{1}\cdot \mathcal{L}_{pos}^{m1,m2}+{\lambda}_{2}\cdot \mathcal{L}_{neg}^{m1,m2}
\end{equation}
\end{small}
where \(m1, m2\) denote specified two modalities, \({\lambda}_{1}, {\lambda}_{2}\) are two hyperparameters, \(\mathcal{L}_{pos}\) denotes the loss of positive samples, it is used to bring features of different instances with the same id closer, and \(\mathcal{L}_{neg}\) denotes the loss of negative samples, it is designed to bring the features of different ids farther. Its specific definition is as follows:
\begin{small}
\begin{align}
\mathcal{L}_{pos}^{m1,m2}&=\frac{1}{P}\sum_{i=1}^PD_{ii}^{m1,m2} \\
\mathcal{L}_{neg}^{m1,m2}&=\frac{1}{P(P-1)}\sum_{i=1}^P\sum_{j=1}^P\frac{1}{D_{ij}^{m1,m2}+\epsilon}, i \neq j
\end{align}
\end{small}
in which \(P\) denotes the number of different people in a training batch. $\epsilon$ is a non-zero constant.

\label{mtc} The \textbf{Modality-Transition Contrastive Constraints} $\mathcal{L}_{mtc}$ is designed to pull the feature distance between the same person in different modalities closer and push the feature between different person away. It exploits the generated infrared image as an intermediate modality to reduce the feature differences between visible and infrared modalities, which consists of three modality-specific constraints:
\begin{small}
\begin{eqnarray}
\mathcal{L}_{mtc}=(\mathcal{L}_{mc}^{v,i} + \mathcal{L}_{mc}^{v,g} + \mathcal{L}_{mc}^{i,g})/3
\end{eqnarray}
\end{small}

$\mathcal{L}_{mc}^{v,i}$ is a regular cross-modal constraint to draw the visible and infrared modalities closer and farther for instances with the same ID and different IDs for each of the two modalities. However, considering that it is impossible for visible and infrared features to be identical, that means, as training increases, it will reach its limits. Therefore we introduce the generated auxiliary modes on this basis to help the model learn further, i.e. $\mathcal{L}_{mc}^{v,g}$ and $\mathcal{L}_{mc}^{i,g}$. 

Since the semantic information is fully aligned in the instances from visible and generated IR images, $\mathcal{L}_{mc}^{v,g}$ helps the model to better correlate the same IDs in two modalities and enables efficient inter-modal alignment due to the high degree of consistency between images. In addition, it also brings visible ID centers closer to infrared ID centers.
$\mathcal{L}_{mc}^{i,g}$ conducts cross-modal spectral constraint. This utilizes the similarity between modal spectra. While constraining internal feature centers of the infrared modality, features are brought closer to corresponding visible ID centers. 
\\
\textbf{Global Center Loss.}
$\mathcal{L}_{mtc}$ is performed only in one training batch. We assist it in our model not only constrains within a batch, but also aligns features globally. As shown in Fig. \ref{fig:main}, the implicit batch center of each ID will be drawn towards this global feature center to improve the model's generalization ability by optimizing the center loss for the entire batch. We use Euclidean distance to compute the distance between \(f_i\) and its class center \(c_i\) as followings.
\begin{small}
\begin{equation}
\mathcal{L}_{center} = \frac{1}{B}\sum_{i=1}^B\Vert f_i - c_i \Vert_2 \label{center_loss}
\end{equation}
\end{small}
where $B$ denotes the batch size of the training samples, which is $3\times N$, and $N$ is the number of positive samples. Then, we average the center distances of all samples in the same batch to obtain the final loss value.

During back-propagation stage, both the center positions \(c_i\) and the sample features \(f_i\) are optimized accordingly, enhancing within-class features and separated between-class features. Overall, the Modality-Aware Hierarchical Constraints (Modality-Aware Hierarchical Constraints is:
\begin{small}
\begin{equation}
\mathcal{L}_{mhc} = \alpha \cdot \mathcal{L}_{mtc} + \beta \cdot \mathcal{L}_{center}
\end{equation}
\end{small}

\begin{table*}
\small
\centering
\renewcommand{\arraystretch}{1}
\renewcommand\tabcolsep{5pt}
             \caption{Comparison with state-of-the-art methods on SYSU-MM01. $^R$ means using Re-Rank. Both \textbf{gray} and \textbf{CycleGan generated} transition image has been compared, which shows even using the gray images as transition modality (\textbf{zero extra cost}) gray still has great improvements.}

	\begin{tabular}{l|c|cc|cc|cc|cc}
		\hline
		\multirow{3}{*}{Model} &\multirow{3}{*}{Venue}& \multicolumn{4}{c|}{All-Search} &\multicolumn{4}{c}{Indoor-Search}\\ \cline{3-10}
              & &\multicolumn{2}{c|}{ Single-Shot} &\multicolumn{2}{c|}{ Multi-Shot} &\multicolumn{2}{c|}{ Single-Shot} &\multicolumn{2}{c}{ Multi-Shot}\\ \cline{3-10}
              & & mAP & Rank-1 & mAP & Rank-1 & mAP & Rank-1 & mAP & Rank-1 \\ \cline{1-10}
             NFS~\cite{9577446} &CVPR'21& 55.45 &56.91& 48.56 &63.51& 69.79 &62.79& 61.45 &70.03\\
             cm-SSFT~\cite{Lu2020CrossModalityPR}&CVPR'20& 63.20 &61.60& 62.00 &63.40& 72.60 &70.50& 72.40 &73.00\\
             CMTR\cite{Liang2021CrossModalityTW}&TMM'23& 61.33 &62.58& 55.69 &68.39& 73.78 &67.02& 66.84 &75.40\\
             MCLNet~\cite{2021cross}&ICCV'21& 61.98 &65.40& - &-& 76.58 &72.56& - &-\\
             MAUM~\cite{Liu2022LearningMU}&CVPR'22& 68.79 &71.68& - &-& 81.94 &76.97& - &-\\
             CAL\cite{CAL}&ICCV'23& 71.73 &74.66& 64.86 &77.05& 83.68 &79.69& 78.51 &86.97\\
             SAAI~\cite{10377129}&ICCV'23& 77.03 &75.90& 82.39 &82.86& 88.01 &83.20& 91.30 &90.73\\
             SEFL\cite{10205002}&CVPR'23& 72.33 &77.12& - &-& 82.95 &82.07& - &-\\
             PartMix\cite{kim2023partmix}&CVPR'23& 74.62 &77.78& 69.84 &80.53& 84.38 &81.52& 79.95 &87.99\\
             
             MID~\cite{Huang2022ModalityAdaptiveMA} &AAAI'22& 59.40 &60.27& - & - & 70.12 &64.86& - & -\\
             FMCNet~\cite{zhang2022fmcnet}&CVPR'22& 62.51 &66.34& 56.06 &73.44& 74.09 &68.15& 63.82 &78.86\\
             MPANet~\cite{9577837}&CVPR'21& 68.24 &70.58& 62.91 &75.58& 80.95 &76.74& 75.11 &84.22\\
             CMT~\cite{jiang2022cross}&ECCV'22& 68.57 &71.88& 63.13 &80.23& 79.91 &76.90& 74.11 &84.87\\
             protoHPE~\cite{Zhang2023ProtoHPEPH}&ACMMM'23& 70.59 &71.92& - &-& 81.31 &77.81& - &-\\
             MUN~\cite{10377404}&ICCV'23& 73.81 &76.24& - &-& 82.06 &79.42& - &-\\
             MSCLNet~\cite{zhang2022modality}&ECCV'22& 71.64 &76.99& - &-& 81.17 &78.49& - &-\\
             DEEN~\cite{Zhang2023DiverseEE}&CVPR'23& 71.80 &74.70& - &-& 83.30 &80.30& - &-\\
             ARGN~\cite{10814927}&TMM'25& 72.71 &77.04& 68.32 &\textbf{84.32}& 85.26 &83.20& 80.73 &91.05\\
             AMML~\cite{zhang2025adaptive}&IJCV'25& 74.8 &\textbf{77.8}& - &-& 88.3 &88.6& - &-\\
             \cline{1-10}\rowcolor[HTML]{EAFAF1} 
             $\text{MTRL}_{gray}$&-& 76.25 &75.17& 81.66 &82.03& 87.72 &83.47& 90.38 &91.22\\
            \cline{1-10}\rowcolor[HTML]{EAFAF1} 
             $\text{MTRL}_{CycleGan}$&-& \textbf{78.36} &76.80& \textbf{83.22} &83.78& \textbf{89.44} &85.53& \textbf{92.44} &92.06\\
             \cline{1-10}
             CIFT~\cite{Li2022CounterfactualIF}$^R$&ECCV'22& 74.79 &74.08& 75.56 &79.74& 85.61 &81.82& 86.42 &88.32\\
             IDKL\cite{Ren2024ImplicitDK}$^R$&CVPR'24& 79.85 &81.42& 78.22 &84.34& 89.37 &87.14& 88.75 &94.30\\
             \cline{1-10}\rowcolor[HTML]{EAFAF1} 
            $\text{MTRL}_{CycleGan}$$^R$&-& \textbf{84.54} &\textbf{85.51}& \textbf{80.91} &\textbf{86.94}& \textbf{92.61} &\textbf{91.55}& \textbf{90.66} &\textbf{93.56}\\
             \cline{1-10}
	\end{tabular}
             \label{tab:sys}
                              
\end{table*}

\begin{table}
\small
    \centering
    \renewcommand{\arraystretch}{1.1}
    \renewcommand\tabcolsep{5pt}
              \caption{Comparison with SOTA models on RegDB. $^R$ means using Re-Rank.}
    
    \begin{tabular}{c|c|c|c|c}
        \hline
         \multirow{2}{*}{Model}& \multicolumn{2}{c|}{Visible2Infrared} &\multicolumn{2}{c}{Infrared2Visible} \\ \cline{2-5}
          & mAP & Rank-1 & mAP & Rank-1 \\ \cline{1-5}
          cm-SSFT & 72.90 & 72.30 & 71.70 & 71.00 \\
          MCLNet & 73.07 & 80.31 & 69.49 & 75.93 \\
          NFS & 72.10 & 80.54 & 69.79 & 77.95 \\
          MPANet & 80.90 & 83.70 & 80.70 & 82.80 \\
          MSCLNet & 80.99 & 84.17 & 78.31 & 83.86 \\
          MID & 84.85 & 87.45 & 81.41 & 84.29 \\
          MAUM & 85.09 & 87.87 & 84.34 & 86.95 \\
          FMCNet & 84.43 & 89.12 & 83.86 & 88.38 \\
          SAAI & 91.45 & 91.07 & 92.01 & 92.09 \\
          CMT & 87.30 & 95.17 & 84.46 & 91.97 \\
          MUN & 87.15 & 95.19 & 85.01 & 91.86 \\
          DEEN & 85.10 & 91.10 & 83.4 & 89.5 \\
          ARGN & 90.02 & \textbf{96.16} & 87.83 & 94.14 \\
          AMML & 87.8 & 94.9 & 86.3 & 92.1 \\
          \cline{1-5}\rowcolor[HTML]{EAFAF1} 
          $\text{MTRL}_{gray}$ & 91.17 & 92.40 & 90.65 & 91.45 \\
          \cline{1-5}\rowcolor[HTML]{EAFAF1} 
          $\text{MTRL}_{CycleGan}$ & \textbf{94.18} & 93.64 & \textbf{93.98} & \textbf{93.54} \\
          \cline{1-5}
          CIFT$^R$ & 92.00 & 91.96 & 90.78 & 90.30 \\
          IDKL$^R$ & 90.19 & 94.72 & 90.43 & 94.22 \\
          \cline{1-5}\rowcolor[HTML]{EAFAF1} 
          $\text{MTRL}_{CycleGan}$$^R$ & \textbf{96.29} & \textbf{98.01} & \textbf{96.04} & \textbf{97.96} \\
          \cline{1-5}
          \end{tabular}
          \label{tab:regdb}    

\end{table}

\begin{table}
\small
    \centering
    \renewcommand{\arraystretch}{1.1}
    \renewcommand\tabcolsep{5pt}
    \caption{Comparison with SOTA models on LLCM dataset. $^R$ means using Re-Rank.}
    \begin{tabular}{c|c|c|c|c}
        \hline
         \multirow{2}{*}{Model}& \multicolumn{2}{c|}{Visible2Infrared} &\multicolumn{2}{c}{Infrared2Visible} \\ \cline{2-5}
          & mAP & Rank-1 & mAP & Rank-1 \\ \cline{1-5}
          DDAG\cite{Ye2020DynamicDA} & 48.4 & 40.3 & 52.3 & 48.0 \\
          CAJ\cite{ye2021channel} & 59.8 & 56.5 & 56.6 & 48.8 \\
          DEEN\cite{Zhang2023DiverseEE} & 65.8 & 62.5 & 62.9 & 54.9 \\
          ARGN\cite{10814927} & \textbf{66.6} & 63.9 & \textbf{63.3}& 56.9 \\
          AMML\cite{zhang2025adaptive} &53.4&68.3&60.8&54.3 \\
           \cline{1-5}\rowcolor[HTML]{EAFAF1} 
          $\text{MTRL}_{CycleGan}$ & 65.25& \textbf{71.28}  & 59.60 & \textbf{66.26} \\ 
          \cline{1-5}\rowcolor[HTML]{EAFAF1} 
          $\text{MTRL}_{CycleGan}$$^R$& 65.11 & 74.84  & 62.90 & 69.68 \\
          \cline{1-5}
          \end{tabular}
          \label{tab:llcm}    
\end{table}

\subsection{Modality-Query Regularization Mechanism}

To further regularize differences between modalities, we take one modality as a query for another modality and employ the generated modality for better feature alignment. Besides, we make full use of the ID distance matrix obtained by calculating Modality-Transition Hierarchical Constraints loss to improve efficiency as declared in Fig.~\ref{mtc}. As Fig.~\ref{fig:main}, each matrix of positive samples contains distance information under visible, generated infrared, and infrared modals. Besides, by studying the distance matrix of positive instances, where each small matrix $E$ represents the average distance of the same person between two modalities. This can be abstracted as an inter-modal distance query operation. For example, each row in \(E^{vi}\) represents the distance of a particular visible instance (\(v\)) querying for infrared instances (\(i\)). 

Based on these distance queries, we proposed \textbf{Modality-Query Regularization} $\mathcal{L}_{mqr}$ as:
\begin{small}
\begin{align}
\mathcal{L}_{mqr}&=(\mathcal{L}_{qr}^{vg,vi}+\mathcal{L}_{qr}^{gi,vi}+ \mathcal{L}_{qr}^{ig,iv}+\mathcal{L}_{qr}^{gv,iv})/4 \\
\mathcal{L}_{qr}^{a,b}&=\frac{1}{N}\sum_{i=1}^N{\left\lVert \sum_{j=1}^NE^{a}_{ij} - \sum_{j=1}^NE^{b}_{ij} \right\rVert_2} 
\end{align}
\end{small}
where $\mathcal{L}_{qr}^{vg,vi}$ and $\mathcal{L}_{qr}^{gv,iv}$ aim to reduce the difference between the visible-to-infrared query distance while $\mathcal{L}_{qr}^{ig,iv}$ and $\mathcal{L}_{qr}^{gi,vi}$ are used to reduce the difference between the infrared-to-visible query distance. The principle is similar to $\mathcal{L}_{qr}^{vg,vi}$ and $\mathcal{L}_{qr}^{gv,iv}$ above. \(N\) is the number of positive samples, \(E\) is a distance matrix mentioned in Fig. \ref{mtc} which only contains positive samples distances between two modalities, as shown in Fig. \ref{fig:main}. $a, b$ refers to two query matrices. Specifically, $E^{a}_{ij}$ denotes the distance of the $i_{th}$ instance from the $j_{th}$ instance of matrix $E^a$.

Specifically, For $\mathcal{L}_{qr}^{vg,vi}$, since visble and generated IR modalities are fully aligned, we use each visble instance to query for the distance between aligned generated-IR instances $E^{vg}$ and real-IR intances $E^{vi}$ respectively, and using $E^{vg}$ to guide the model to make the visble-to-infrared query difference smaller. For $\mathcal{L}_{qr}^{gi,vi}$, since there is a significant modal similarity between generated IR and IR modality in terms of spectra, we use each generated-IR instance and visble instance to query for the distance with IR instances to get $E^{gi}$ and $E^{vi}$, and using $E^{gi}$ to guide the model to make the visble-to-infrared query difference smaller.

\subsection{Training and Inference}

During training, we utilize pre-generated IR-transition images to pass through the backbone along with visible and real IR modalities to extract features, and optimize them using the following loss:
\begin{small}
\begin{equation}
\mathcal{L}=\mathcal{L}_{id}+\underbrace{\alpha\cdot\mathcal{L}_{mtc}+\beta\cdot\mathcal{L}_{center}}_{\mathcal{L}_{mhc}}+\gamma\cdot\mathcal{L}_{mqr}
\end{equation}
\end{small}
where $\alpha$, $\beta$, and $\gamma$ are three hyper-parameters for balance. $\mathcal{L}_{id}$ is a classification loss learn identity information, and details are provided in the \textbf{supplementary}.

Particularly, the generative modality does not participate in the inference process, beneficial to achieving commensurate inference efficiency with the typical models.

\section{Experiments}
\subsection{Datasets and Evaluation Metrics}

\textbf{SYSU-MM01}\cite{2017RGB} is the largest benchmark dataset for VI-ReID tasks, containing 286,628 visible and 15,792 infrared images across 491 unique identities. This dataset is split into a training set with 395 persons (22,258 visible and 11,909 infrared images) and a testing set comprising 96 persons with 3,803 infrared images for queries and a gallery of 301 visible images. It introduces two modes: \textit{all-search mode}, which includes all images in the testing phase, and \textit{indoor-search mode} focusing solely on images captured indoors. For both modes, we utilize the most challenging \textit{single-shot} setting for evaluation. The tests are conducted 10 times to obtain the mean performance metrics.

\textbf{RegDB}\cite{s17030605} consists of 412 unique identities with a total of 8,240 images, in which half of identities are for training and the others are for testing. Besides, each identity has 10 visible and 10 infrared images. The evaluation on RegDB includes two patterns: (1) Visible-to-infrared (V2I), which means to match infrared images with their visible counterparts. (2) Infrared-to-visible (I2V), where the objective is identifying visible images with infrared queries.

\textbf{LLCM}\cite{Zhang2023DiverseEE} dataset is a large-scale and low-light cross-modality dataset, which is divided into training and testing sets at a 2:1 ratio.Both LLCM and RegDB contain visible-to-infraed and infrared-to-visible two search modes.

\textbf{Metrics.} We utilize two standard assessment metrics namely Cumulative Matching Characteristic (CMC) and mean Average Precision (mAP). Besides, using k-reciprocal\cite{8099872} re-rank method to compare with those SOTA models using re-rank.

\begin{table}
\small
 \renewcommand{\arraystretch}{1.0}
 \renewcommand\tabcolsep{4pt}
\caption{Ablation study of each proposed module on SYSU-MM01 under the All-Search and Indoor-Search.}
 \begin{tabular}{cccc|cc|cc}
\hline
\multicolumn{4}{c|}{\textbf{Loss}} & \multicolumn{2}{c|}{\textbf{All-Search}} & \multicolumn{2}{c}{\textbf{Indoor-Search}} \\ \hline
$\mathcal{L}_{id}$ & $\mathcal{L}_{mtc}$ & $\mathcal{L}_{mqr}$ & $\mathcal{L}_{center}$ & mAP & R1 & mAP & R1 \\ \hline
\ding{51} &  \ding{55} &\ding{55} & \ding{55} & 73.57 & 72.79 & 85.23 & 79.97 \\
\ding{51} & \ding{51} & \ding{55} & \ding{55} & 76.66 & 75.94 & 88.56 & 84.45 \\
\ding{51} & \ding{55} & \ding{55} & \ding{51} & 74.41 & 73.95 & 86.02 & 80.68 \\
\ding{51} & \ding{55} & \ding{51} & \ding{55} & 74.67 & 74.23 & 86.16 & 81.32 \\
\ding{51} & \ding{51} & \ding{51} & \ding{55} & 77.17 & 76.01 & 88.66 & 84.62 \\
\ding{51}  & \ding{51}  & \ding{51} & \ding{51} & \textbf{78.36} & \textbf{76.80}  & \textbf{89.44} & \textbf{85.53} \\ \hline
\end{tabular}
\label{tab:loss}
\end{table}

\begin{figure}
\centering
\includegraphics[width=\linewidth]{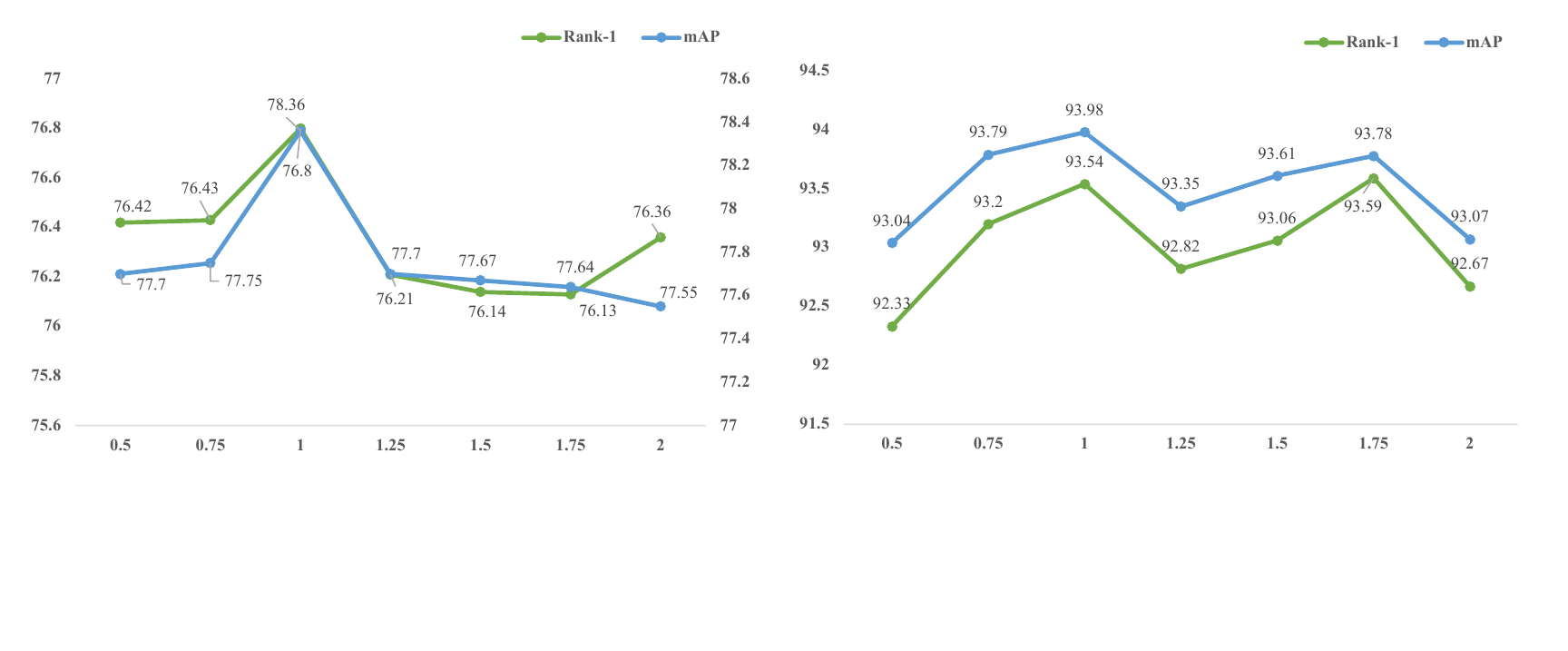}
\caption{Sensitive graph of \textbf{$\alpha$} on SYSU-MM01 with All-Search and Single-Shot settings, and RegDB with I2V mode.}
\label{fig:alpha}
\end{figure}
\begin{figure}
\centering
\includegraphics[width=\linewidth]{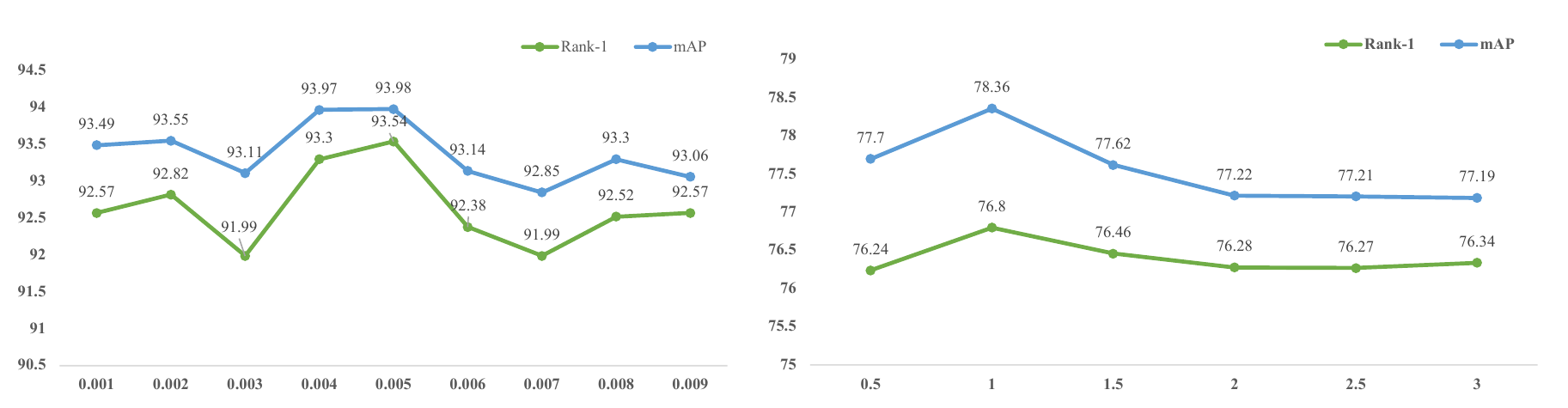}
\caption{Sensitive graph of \textbf{$\beta$} (left) on RegDB with I2V mode and \textbf{$\gamma$} (right) on SYSU-MM01 with All-Search and Single-Shot mode. }
\label{fig:gamma}
\end{figure}
\begin{figure}
\centering
\includegraphics[width=0.9\linewidth]{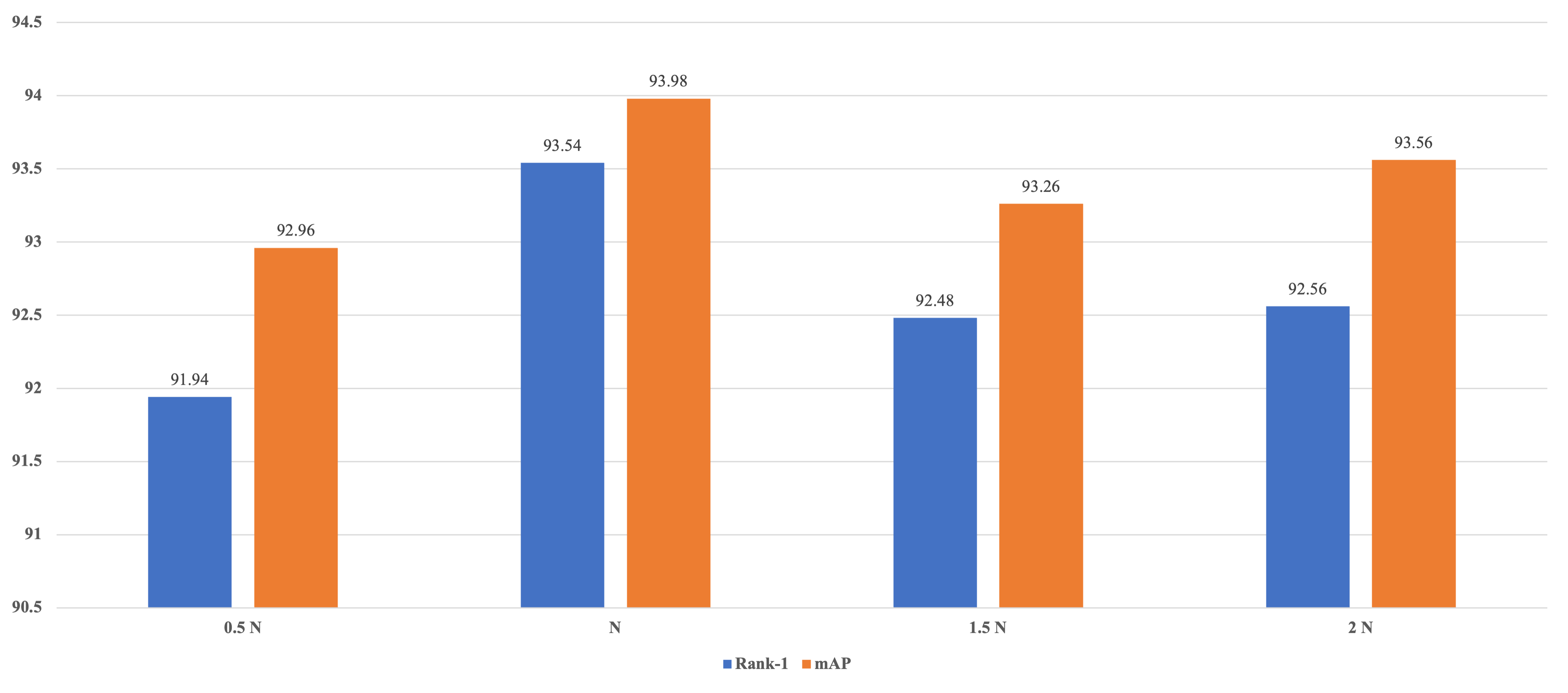}
\caption{The impact of the number of top-k on RegDB with I2V mode. $N$ is the number of positive samples.}
\label{fig:topk_regdb}
\end{figure}

\begin{table}
    \centering
    \renewcommand{\arraystretch}{1.1}
    \renewcommand\tabcolsep{5pt}
    \caption{Comparison on baselines with/without our framework.}
    \begin{tabular}{c|c|c|c|c}
        \hline
        \multirow{2}{*}{Baseline} & \multicolumn{2}{c|}{All-Search} & \multicolumn{2}{c}{Single-Search} \\ \cline{2-5}
        & Rank-1 & mAP & Rank-1 & mAP \\ \cline{1-5}
        ResNet50 & 68.24 & 69.42 & 74.49 & 81.04 \\
        \textbf{ResNet50+MTRL} & \textbf{70.76} & \textbf{72.98} & \textbf{79.18} & \textbf{84.94} \\
        ResNet18 & 54.53 & 56.07 & 59.42 & 68.85 \\
        \textbf{ResNet18+MTRL} & \textbf{56.06} & \textbf{59.54} & \textbf{61.84} & \textbf{71.7} \\
        \hline
    \end{tabular} 
    \label{res}
\end{table}

\subsection{Implementation Details}

We conduct experiments with PyTorch and an NVIDIA RTX-3090 GPU. The backbone follows the settings of SAAI~\cite{10377129}, which utilizes ResNet-50 to extract a global and 7 part-features. To ensure reproducibility and fair comparisons with existing models, we employ the official pre-trained model for ResNet-50.

Input images are initially resized to a consistent dimension of $288\times144$ and a series of augmentation techniques have been applied, including random cropping, random erasing, and random horizontal flipping. For each batch, we randomly sample 16 identities and each identity contains 4 positive images for SYSU-MM01, and 8 identities and each identity contains 2 positive images for RegDB since this dataset is relatively small. 
The network is optimized by Adam with a linear warm-up strategy. The initial learning rate is set to $3.5 \times 10^{-4}$ and is decreased by factors of 0.1 and 0.01 at 80 and 180 epochs, respectively. The training procedure spans a total of 250 epochs. The top-k number $k$ in $\mathcal{L}_{mtc}$ is set to $N$ (the positive samples). The two hyper-parameters \({\lambda}_{1}, {\lambda}_{2}\) of $\mathcal{L}_{mc}$ are set to 1.0 and 0.1. The loss balance hyper-parameters $\alpha$, $\beta$, and $\gamma$ are 1.0, 0.005, 1.0.

\begin{figure*}
\centering
\includegraphics[width=0.9\linewidth]{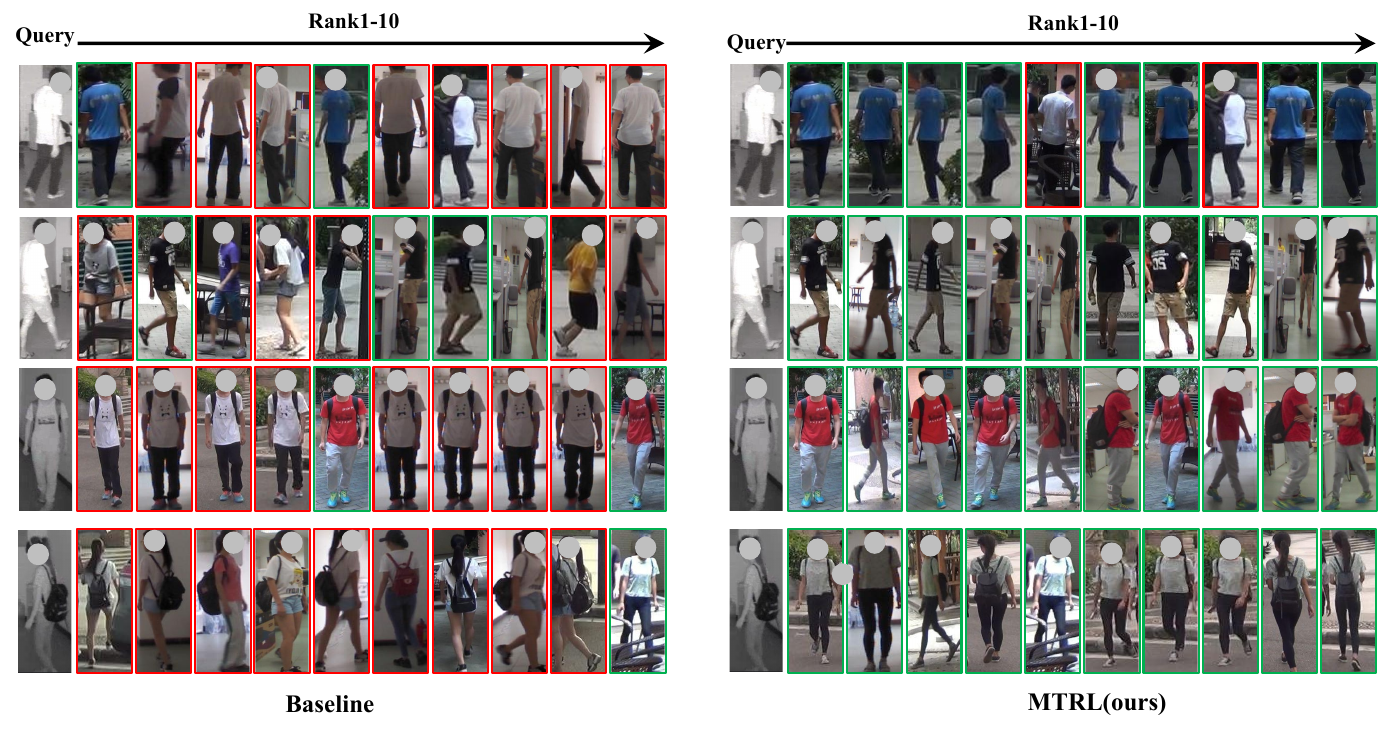}
\caption{Illustration of person retrieval result on SYSU-MM01. The left is baseline, and the right is our MTRL.}
\label{fig:rank10}
\end{figure*}

\begin{figure}
\centering
\includegraphics[width=\linewidth]{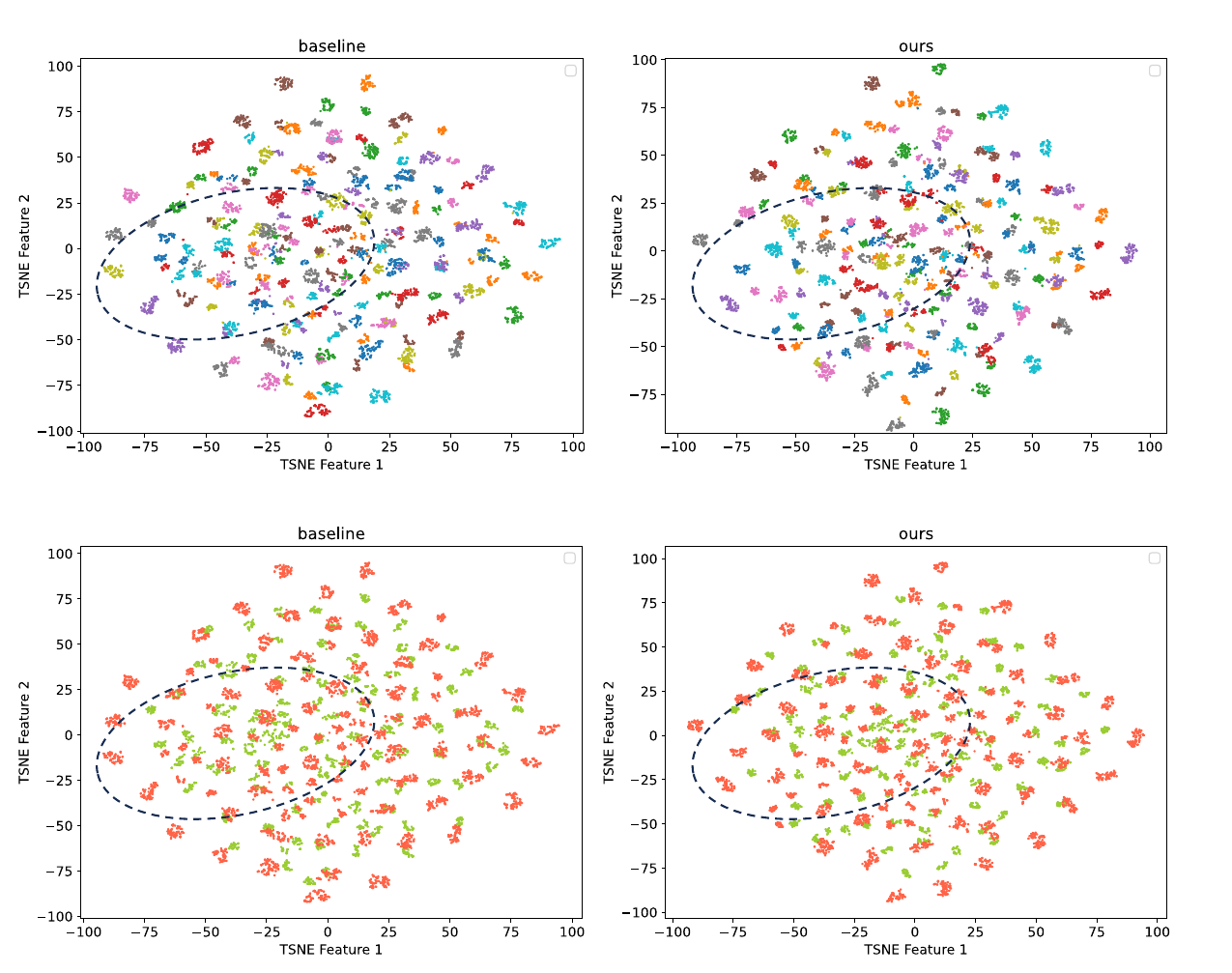}
\caption{Features distribution of baseline and ours on SYSU-MM01. First row: colors represent identities; Second row: colors represent modalities}
\label{fig:point}
\end{figure}

\subsection{Comparison with State-of-the-Art Methods}

We compare the proposed approach MTRL with some classical and advanced state-of-the-art (SOTA) methods on both SYSU-MM01, RegDB and LLCM datasets. The results are shown in Table~\ref{tab:sys}, Table\ref{tab:regdb} and Table~\ref{tab:llcm}. The best performing results are bold. 

\textbf{Comparison Results on SYSU-MM01.}
Our model demonstrates superior performance on SYSU-MM01. As observed in Table\ref{tab:sys}, our model MTRL significantly outperforms the SOTA methods across multiple metrics. Notably, without using re-rank, in the All-Search Single-Shot setting, it achieves a remarkable Rank-1 of 76.8\% and mAP of 78.36\%. Similarly, in the Indoor-Search Single-Shot mode, our model achieves an impressive Rank-1 of 85.53\% and mAP of 89.44\%, surpassing the second-best(SAAI) reported Rank-1 by 1.5\% and mAP by 0.71\%. With using re-rank, our model surpasses latest SOTA model IDKL with an average 2.86\% on 8 metrics.

\textbf{Comparison Results on RegDB.}
Our model achieves Rank-1 of 93.54\% and mAP of 93.98\% on I2V retrieval, and Rank-1 of 93.64\% and mAP of 94.18\% on V2I retrieval. Meanwhile, using re-rank, it has an average improvement of 4.68\% compared with latest SOTA model(IDKL).

\textbf{Comparison Results on LLCM.}
Our model achieves Rank-1 of 66.26\% and mAP of 59.60\% on I2V retrieval, and Rank-1 of 71.28\% and mAP of 65.25\% on V2I retrieval. This indicates that our model exhibits strong robustness in complex and multimodal scenarios.

In summary, our MTRL framework outperforms in seven out of eight metrics on SYSU-MM01, in three out of four metrics on RegDB and in three out of four metrics on LLCM without re-rank, illustrating the superiority of our proposed model. Moreover, considering the properties of k-reciprocal re-rank, a significant improvement using re-rank shows the outstanding performance of mutual retrieval between two modalities.

\subsection{Parameters Analysis}

\textbf{Parameters Analysis of $\alpha$.}
We evaluate the influence of the hyper-parameter $\alpha$ on SYSU-MM01 and RegDB as to the all-search and single-shot settings. Fig.~\ref{fig:alpha} shows the results of Rank-1 and mAP of different $\alpha$. The most suitable parameter setting is 1.0.

\textbf{Parameters Analysis of $\beta$ and $\gamma$. }
We evaluate the impact of $\beta$ on RegDB and the impact of $\gamma$ on SYSU-MM01. The most suitable parameter setting for $\beta$ is 0.005 and for $\gamma$ is 1.0, as shown in Fig.~\ref{fig:gamma}.

\textbf{Parameters Analysis of $k$. }
We evaluate the number $k$ in top-k of $\mathcal{L}_{mcc}$ on RegDB with Infrared2Visible mode. As shown in Fig.~\ref{fig:topk_regdb}, the optimal performance is achieved when top-k is set to $N$, i.e., the number of positive samples.

\begin{figure}
\centering
\includegraphics[width=0.9\linewidth]{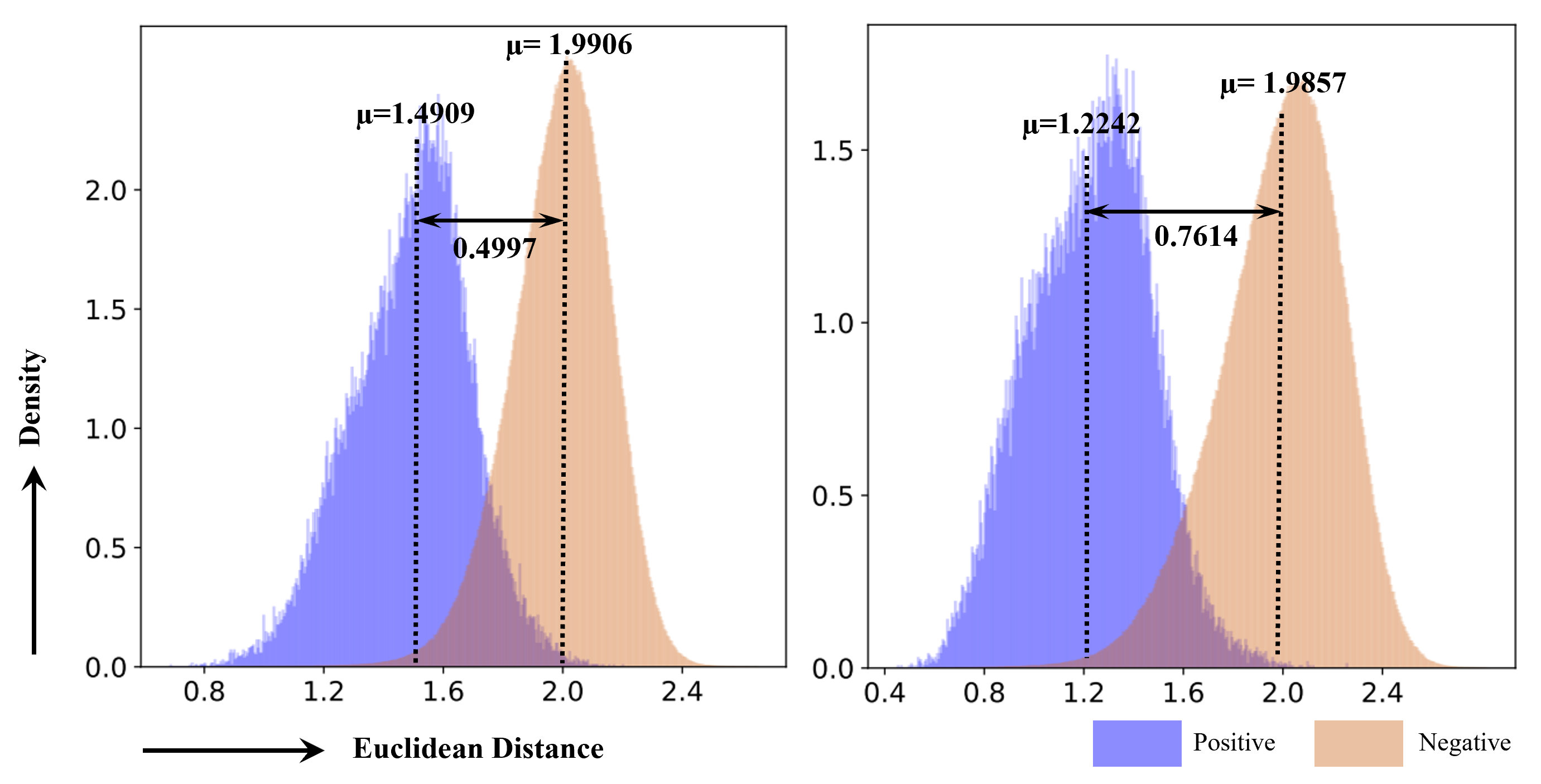}
\caption{Euclidean distance distribution between query and positive/negative samples on SYSU-MM01.}
\label{fig:pos_neg}
\end{figure}

\begin{figure}[]
  \centering
  \includegraphics[width=0.9\linewidth]{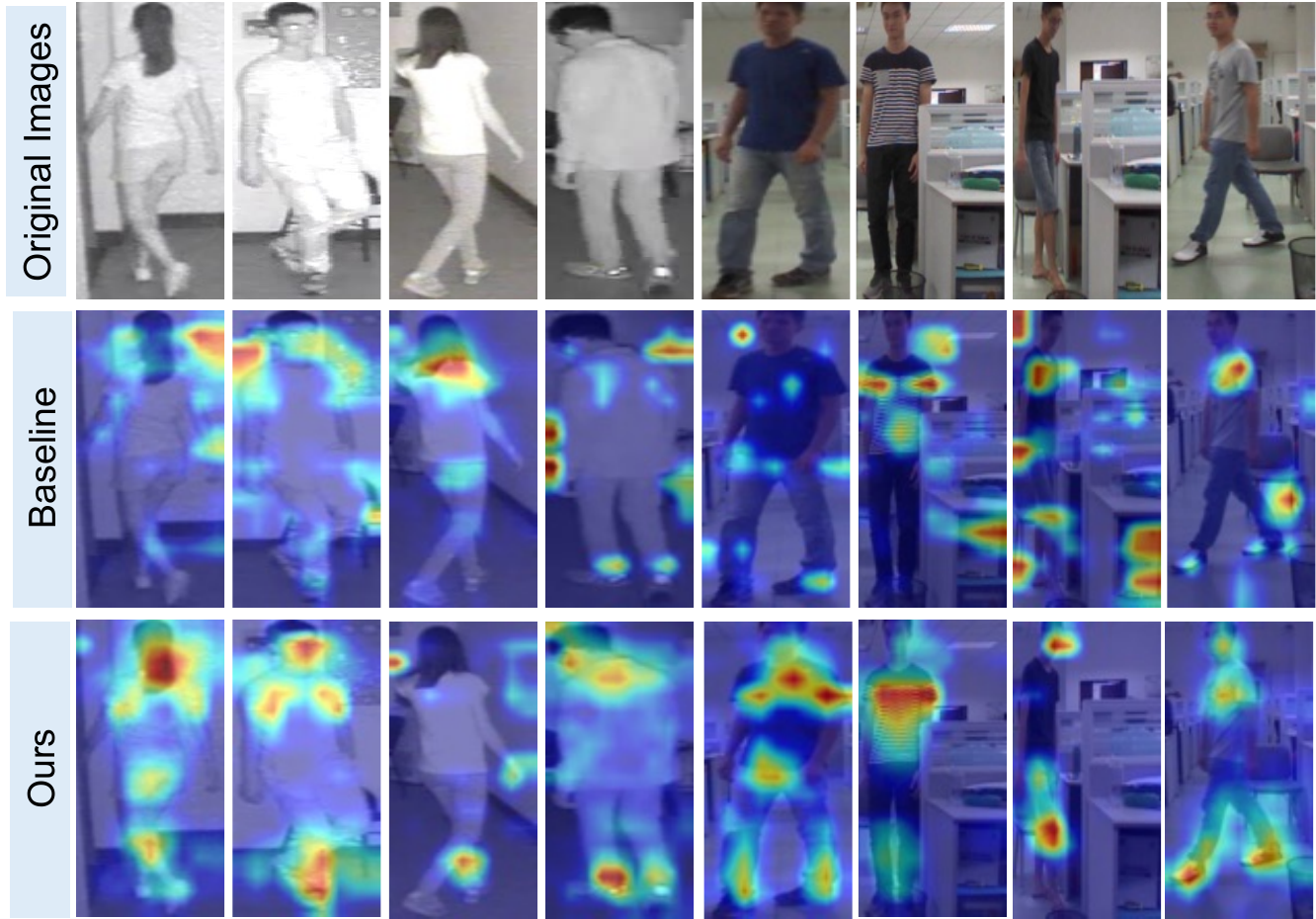}
  \caption{
   Comparisons in Grad-CAM visualization between baseline and our proposed MTRL. 
  }
  \label{heatmap}
\end{figure}

\subsection{Ablation Study}
\textbf{Losses:}
We take the model of SAAI trained by $\mathcal{L}_{id}$ as our baseline. As shown in Table\ref{tab:loss}, we verify effects of each loss function. It improves Rank-1/mAP from 72.79\%/73.57\% to 76.8\%/78.36\%. By adding losses one by one, Rank-1/mAP is improved 4.01\%/4.79\% on All and Single-Shot settings. In detail, $\mathcal{L}_{mhc}$ ($\mathcal{L}_{mtc}$ and $\mathcal{L}_{center}$) enhances Rank-1/mAP by 3.87\%/3.96\%, $\mathcal{L}_{mqr}$ enhances Rank-1/mAP by 1.44\%/1.1\%. It is worth noting that improvements here are without any additional module (parameters). \\
\textbf{MTRL on Other Baselines: }
To verify the portability of our approach, we tested it on ResNet. mAP increases from 69.42 to 72.98 on ResNet50 and from 56.07 to 59.54 on ResNet18, as shown in Table~\ref{res}.\\

\subsection{Visualization}

\textbf{Feature Distribution Analysis.}
We utilize t-SNE to visualize the feature distribution of baseline and MTRL. Fig.\ref{fig:point} illustrates that, compared to the baseline, our loss constraint more effectively separates features that, despite having different IDs, were previously intermingled, and enhances their clustering. The distribution of RGB and IR modality features in the test set is also visualized, showing that features from the same modality and ID are closely clustered, indicating minimal intra-ID variation and pronounced inter-ID differences.\\

\textbf{Retrieval visualization analysis.} 
We visualize the top-5 retrieval results of baseline and MTRL on SYSU-MM01 dataset. The green color indicates the same ID as the query instance and the red color indicates misidentified targets. As shown in Fig.~\ref{fig:rank10}, It can be seen that the baseline pays too much attention to overall features of pedestrians, such as colors. In contrast, ours overcomes this limitation, exhibiting a high accuracy in matching. Due to generative modality-transition learning, model recognizes that colors of clothes are less helpful for cross-modal recognition, and thus actively focuses on specific features. For example, in Fig.~\ref{fig:rank10}, the posture and proportion of the first and second pedestrians, the specific pattern on clothes of the third pedestrian, and the stature and hairstyle of the fourth pedestrian. This clearly illustrates effects of MTRL in accurately matching individuals based on clothing attributes.\\
\textbf{Analysis of Positive and negative samples distribution.} 
We visualizes the distribution of Euclidean distances for features with the same ID (positive samples) versus different IDs (negative samples) in 10 rounds of randomly sampled tests. From Fig.~\ref{fig:pos_neg}, the average distance of MTRL reduces from 1.4909 to 1.2242 compared with baseline. Meanwhile, it increases the mean distance between positive and negative distributions from 0.4997 to 0.7614, which shows that MTRL effectively reduces modality differences.\\
\textbf{Grad-CAM Visualization.}
Fig.~\ref{heatmap} illustrates attention maps generated by our method compared to those of baseline methods. Notably, the baseline often erroneously produce high attentional responses in the background areas of images. In contrast, our method demonstrates a more focused ability to allocate attention to modality-invariant areas within pedestrian images. This enables to extract more compact and robust modality-invariant representations.

\section{Conclusion}
In this paper, we propose a novel generative Modality-Transition Representation Learning framework named MTRL for VI-ReID tasks. It is a training-inference decoupled framework which introduce any extra inference time. Meanwhile, the proposed modality constraints $\mathcal{L}_{mhc}$ and $\mathcal{L}_{mqr}$ could effectively constrain cross-modal person ID center and bring the feature of the same person cross different modalities closer. The empirical analyses are performed comprehensively on SYSU-MM01 and RegDB datasets, demonstrating that our framework MTRL significantly improves the performance on VI-ReID tasks.

{
    \small
    \bibliographystyle{ieeenat_fullname}
    \bibliography{main}
}
\clearpage
\setcounter{page}{1}
\maketitlesupplementary

\section{Identity Loss Explanation}

The identity loss $\mathcal{L}_{id}$ is designed to train the model so that it can accurately classify inputs from different modalities (visible and infrared images) into correct identity classes. It only calculates between the original visible and infrared features ($f^v, f^i$) without the generated compensation modality. All features should be batch-normalized:
\begin{eqnarray}
x^v,x^i=BN(f^v, f^i)
\end{eqnarray}
where $x^v,x^i$ are the features after batch-normalization.

$\mathcal{L}_{id}$ is achieved through several components:

\textbf{Cross-Entropy Loss for General Features:}
The first part of the loss is computed from the general features. These features are passed through a classifier $C(\cdot|\cdot)$ to obtain the logits of $x^{[v,i]}$, and the cross-entropy loss is computed between these logits and the ground truth labels.

\textbf{Modality-Specific Classifiers:}
Features are separated based on their modality. Each set of features is processed by its respective classifier ($C_{v}(\cdot|\cdot)$, and $C_{i}(\cdot|\cdot)$) to obtain modality-specific logits, and cross-entropy losses are computed similarly.

\textbf{Consistency Loss:}
A consistency regularization term is added to ensure that the updated classifiers ($C_{v}^{'}(\cdot|\cdot)$, and $C_{i}^{'}(\cdot|\cdot)$) predict similar distributions as the original classifiers ($C_{v}(\cdot|\cdot)$, and $C_{i}(\cdot|\cdot)$). This is achieved by calculating the cross-entropy loss between the merged logits from the original classifiers and the softmax outputs from the updated classifiers.
\begin{eqnarray}
\mathcal{L}_{id}&=&CE(C(x^{v,i}|\theta^{v,i}), y^{v,i})   \\ \nonumber
&+&CE(C_{v}(x^v|\theta^v), y^v)+CE(C_{i}(x^i|\theta^i), y^i) \\
x^{v,i}&=&ConCat(x^v, x^i) 
\end{eqnarray}
where $CE(\cdot,\cdot)$ denotes Cross-Entropy loss. $C(x|\theta)$ denotes a learnable classifier with weights $\theta$ to classify $x$. $y$ denotes the labels of input $x$.

Then, a simple Exponential Moving Average (EMA) method is used to smooth and update the classifiers’ weights:
\begin{eqnarray}
\tilde{\theta}^v&=&(1 - r) \cdot \tilde{\theta}^v+ r \cdot \theta^v \\
\tilde{\theta}^i&=&(1 - r) \cdot \tilde{\theta}^i+ r \cdot \theta^i 
\end{eqnarray}
where $r$ denotes the update rate, set to 0.2 following SAAI.

Eventually, we get the identity loss $\mathcal{L}_{id}$:
\begin{eqnarray}
z^m&=&ConCat(C_{v}(x^v|\theta^v)), C_{i}(x^i|\theta^i)) \\
\tilde{z}^m&=&ConCat(C_{v}^{'}(x^v|\tilde{\theta}^{i})), C_{i}^{'}(x^i|\tilde{\theta}^{v})) \\
\mathcal{L}_{id}&+=& CE(z^m, Softmax(\tilde{z}^m))
\end{eqnarray}

\begin{table}
\scriptsize
\centering
 \renewcommand{\arraystretch}{1.0}
 \renewcommand\tabcolsep{3pt}
 \caption{Ablation study of Loss $\mathcal{L}_{mtc}$ on SYSU-MM01.}
 \begin{tabular}{c|cc|cc}
\hline
\multirow{2}{*}{Loss $\mathcal{L}_{mtc}$} &\multicolumn{2}{c|}{\textbf{All-Search}} & \multicolumn{2}{c}{\textbf{Indoor-Search}} \\ \cline{2-5} 
 & mAP & R1 & mAP & R1 \\ \hline
$\mathcal{L}_{mc}^{vi}$ & 77.14&	75.80&	88.39&	84.51 \\
$\mathcal{L}_{mc}^{vg}+\mathcal{L}_{mc}^{ig}$ & 77.91&	76.62&	88.86&	85.03 \\
All & \textbf{78.36} & \textbf{76.80}  & \textbf{89.44} & \textbf{85.53} \\ \hline
\end{tabular}
\label{tab:loss Lvi}
\end{table}

\section{Analysis of Modality-Aware Hierarchical Constraints Loss}
The relationship among losses in $\mathcal{L}_{mtc}$ is $ \mathcal{L}_{vi} > \mathcal{L}_{ig} > \mathcal{L}_{vg} $. Loss $ \mathcal{L}_{vi} $ is directly established between two modalities. Due to significant differences between modalities, it may cause instability during early stages of training. As training progresses, model learns a better perception of both modalities, the destabilizing effect of $ \mathcal{L}_{vi} $ diminishes, and can further play a positive role. From Tab.\ref{tab:loss Lvi}, it can be observed that adding $ \mathcal{L}_{vi} $ provides a certain improvement.

\begin{figure}
\centering
\includegraphics[width=0.95\linewidth]{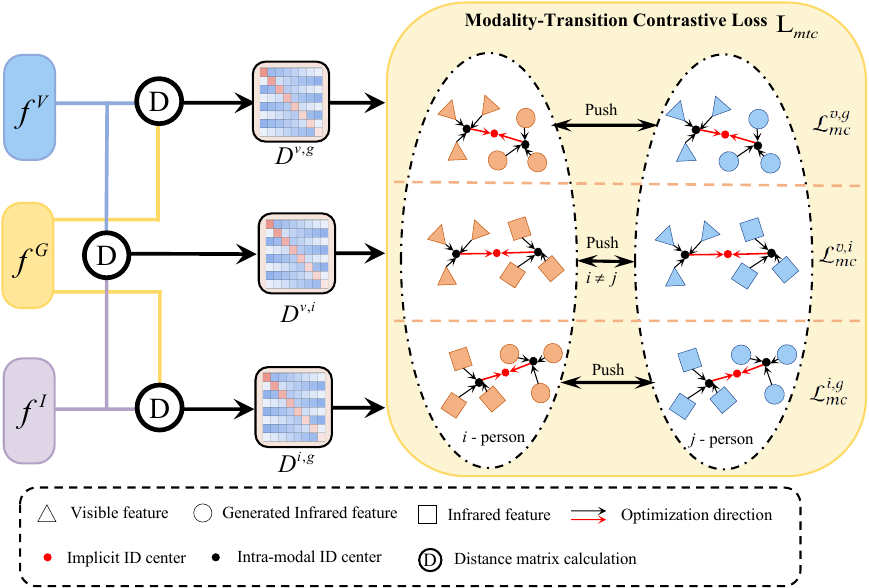}
\caption{Illustration of Modality-Transition Contrastive Loss $\mathcal{L}_{mtc}$ consisting of three losses.}
\label{fig:mtc}
\end{figure}

Moreover, Fig.\ref{fig:mtc} is an illustration of Modality-Transition Contrastive Loss $\mathcal{L}_{mtc}$. Each loss gathers instances from two modalities of the same ID, and pushes different IDs away. In addition, it will implicitly gather instances of two modalities to an implicit ID center, and then global center loss int Eq.(\ref{center_loss}) optimizes it further.

\begin{figure}[]
  \centering
  \includegraphics[width=\linewidth]{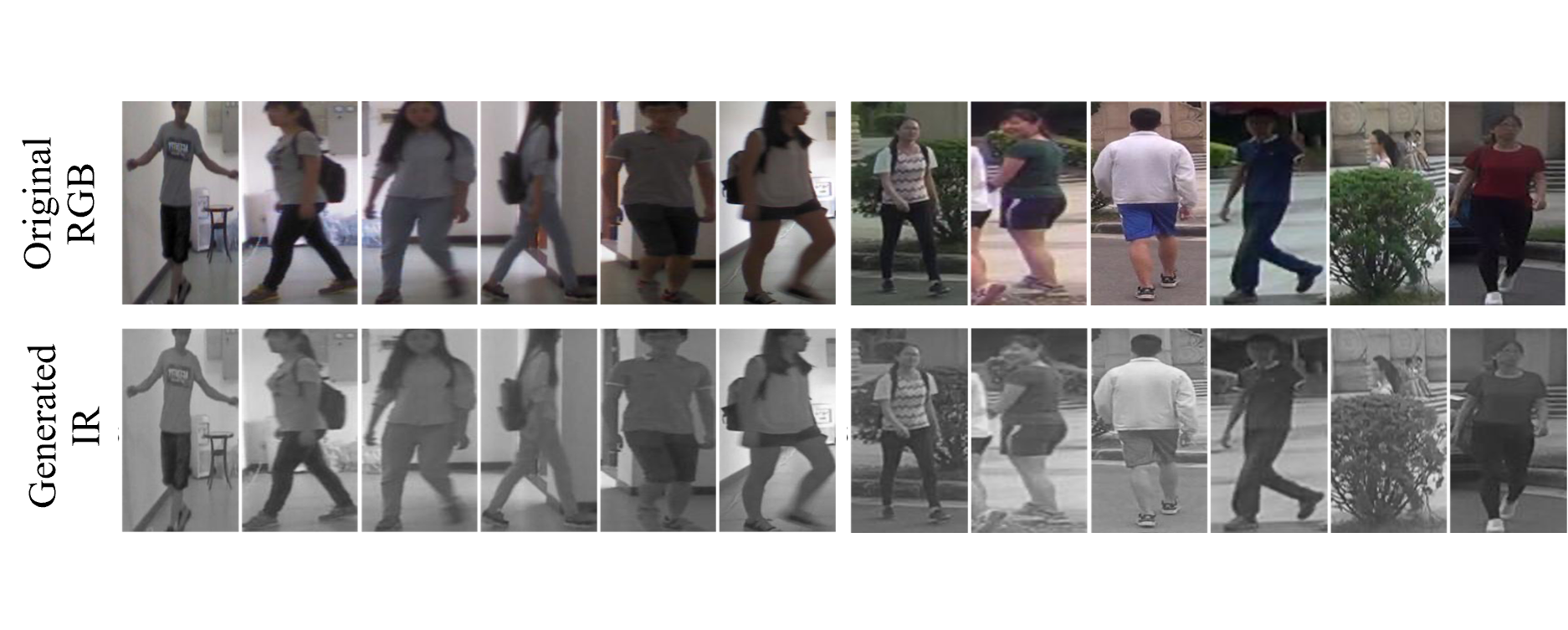}
  \caption{
   Comparison between Generated and Original Images. 
  }
  \label{supgene}
\end{figure}
\section{Visualization of Generation Results.}
Fig.~\ref{supgene} presents a visual comparison between original and generated images.

\end{document}